\documentclass[11pt]{article}

\usepackage[letterpaper,margin=1in]{geometry}
\usepackage[T1]{fontenc}
\usepackage[utf8]{inputenc}
\usepackage{lmodern}
\usepackage{microtype}
\usepackage{amsmath,amssymb,amsthm}
\usepackage{booktabs}
\usepackage{array}
\usepackage{tabularx}
\usepackage{graphicx}
\usepackage{xcolor}
\usepackage{enumitem}
\usepackage{authblk}
\usepackage{hyperref}
\hypersetup{
  colorlinks=true,
  linkcolor=blue!50!black,
  citecolor=blue!50!black,
  urlcolor=blue!60!black,
  pdftitle={Live Gurbani Tracking: A Benchmark and Reference System for Captioning Sikh Kirtan},
  pdfauthor={Karanbir Singh}
}
\usepackage{xurl}
\usepackage[numbers,sort&compress]{natbib}
\usepackage{abstract}

\usepackage{titlesec}
\usepackage{float}
\titlespacing*{\section}{0pt}{1.2ex plus 0.4ex minus 0.2ex}{0.6ex plus 0.2ex}
\titlespacing*{\subsection}{0pt}{0.9ex plus 0.3ex minus 0.2ex}{0.4ex plus 0.2ex}

\usepackage{listings}
\newcommand{\shabad}{\textsf{shabad\_id}}
\newcommand{\lineidx}{\textsf{line\_idx}}

\title{\bfseries Live Gurbani Tracking:\\
A Benchmark and Reference System for Captioning Sikh Kirtan}

\author{Karanbir Singh}
\date{May 2026}

\begin{document}
\maketitle

\begin{abstract}
\noindent
We present a benchmark and reference system for \textbf{live captioning of Sikh Kirtan} - the continuous, sung recitation of verses from the Sri Guru Granth Sahib Ji (SGGS). Unlike open-vocabulary lyrics transcription, Kirtan captioning is a \emph{closed-vocabulary} problem: every displayed line must be an exact, word-for-word line from the canonical scripture, because displaying misspelled Gurmukhi is considered religiously inappropriate. We formalize the task as predicting, at every time $t$, a pair $(\shabad, \lineidx)$ or $\varnothing$, and organize the problem space into a $2 \times 2$ matrix along two orthogonal axes: \textbf{live vs.\ offline} (causal vs.\ full-audio access) and \textbf{blind vs.\ oracle} (shabad identity discovered vs.\ given). We release v1 of the benchmark - 4 hand-annotated Kirtan recordings $\times$ 3 cold-start offsets $= 12$ evaluation cases, ${\sim}57$ minutes of scored audio - together with a scorer that computes frame accuracy at 1\,s resolution over a scored region, with a 1\,s collar and gap-tolerant scoring at segment boundaries. We describe a reference system (fine-tuned 120M IndicConformer $\rightarrow$ fuzzy matcher $\rightarrow$ state machine; INT8 ONNX; RTF ${\approx}0.05$ on one Apple Silicon core) that achieves \textbf{57.9\%} overall frame accuracy across all 12 cases (10/12 correct shabad locks) on the hardest variant (live $\times$ blind). We compare against three trivial baselines (empty, shifted-5\,s, perfect) and discuss why standard ASR metrics (WER/CER) measure transcription accuracy rather than the display accuracy this task requires. The benchmark, reference system, and a live deployment (current demo link in the reference-system repository) are released under permissive licenses to facilitate further improvements.
\end{abstract}


\section{Introduction}

\begin{figure}[h]
\centering
\includegraphics[width=0.85\linewidth]{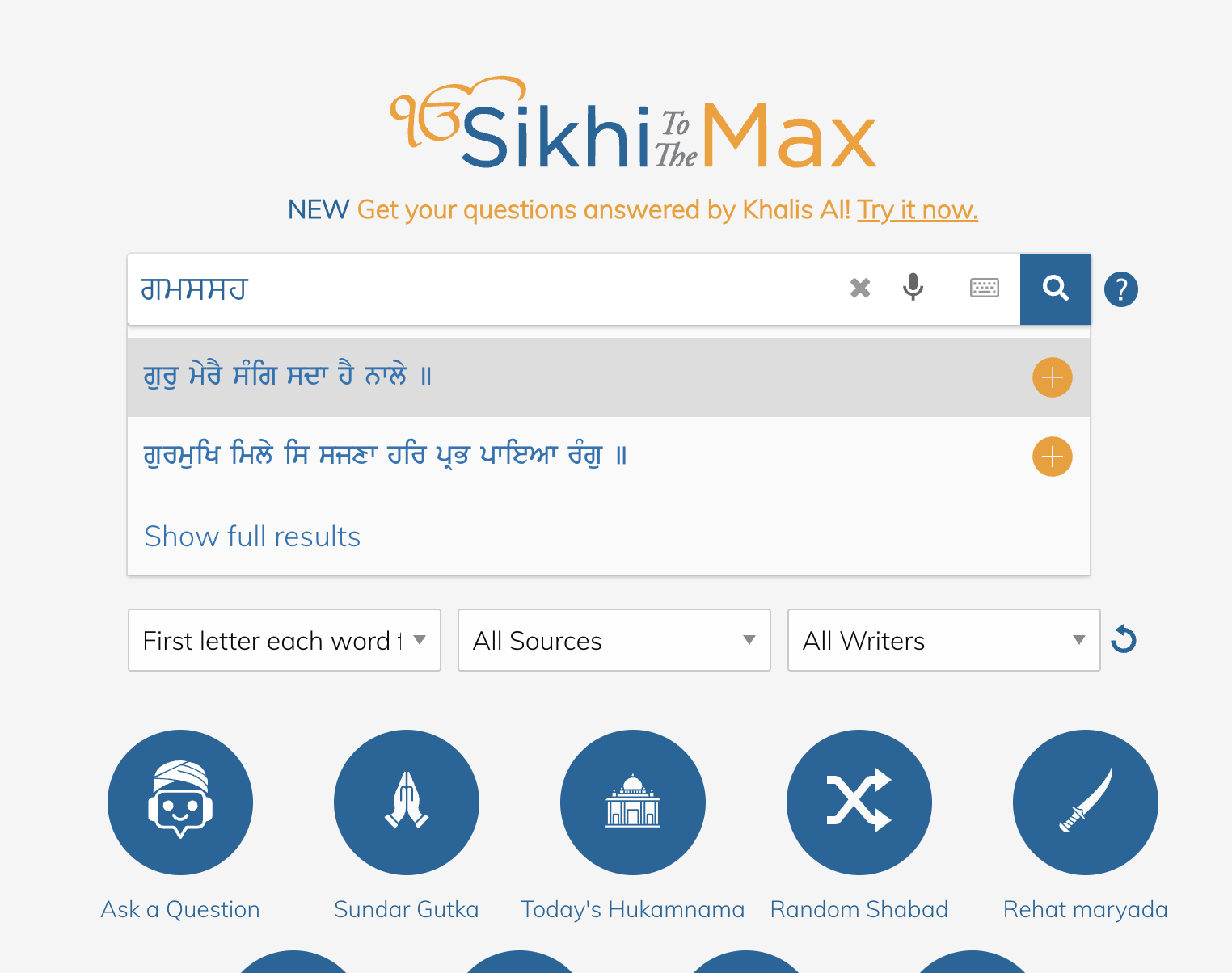}
\caption{Sikhi To The Max (STTM) driving a Gurdwara projector. The sewadar searches by typing the first letter of each word they hear (left), selects the matching shabad from a candidate list, and then clicks line-by-line through the shabad's verses as the singer advances.}
\label{fig:sttm}
\end{figure}

Sikh Kirtan - the devotional singing of verses from the Sri Guru Granth Sahib Ji (SGGS) - is the central spiritual practice of Sikhi. In Gurdwaras (Sikh houses of worship), listeners follow along with a projected display showing the canonical Gurmukhi text of the line currently being sung. Today the display is driven manually by a volunteer (\emph{sewadar}) using tools such as Sikhi To The Max (STTM): the sewadar listens for the opening words of the next shabad, types the first letter of each word into a search box, picks the matching shabad from a candidate list, and then clicks line-by-line through the shabad's verses as the singer advances (Figure~\ref{fig:sttm}).

This end-to-end process is also leveraged by Kirtan video editors to caption recordings. Personal listening applications also offer a way to search and navigate through Gurbani. But across all of these settings the task is not an open-vocabulary lyrics transcription problem. It is instead a \textbf{closed-vocabulary line-identification problem with a hard cultural constraint}: the displayed line must be a verbatim canonical line, because displaying misspelled Gurmukhi is considered religiously inappropriate. Free-form ASR - even very accurate ASR - is therefore not acceptable as the output layer. This is not a hypothetical concern: automatically generated captions already accompany online videos of Kirtan, and where they come from free-form transcription we have observed them rendering incorrect or misspelled Gurmukhi - showing scripture wrongly to every viewer. That failure mode motivates the canonical-output constraint we place at the center of both the task and the metric. Some component, somewhere in the system, must \textbf{snap the output to the canonical text}.

This paper makes four contributions:

\begin{enumerate}[leftmargin=*,itemsep=2pt]
  \item \textbf{The canonical-output constraint as a metric-layer property.} We argue that for Gurbani captioning the religious-appropriateness constraint belongs not just in the system but in the metric: every prediction is checked for an exact match against the canonical Gurbani text, so the scorer rewards only verbatim canonical output or $\varnothing$, never approximate Gurmukhi.
  \item \textbf{A frame-temporal display-accuracy metric} tied to the deployment question - ``is the right line on screen at time $t$?'' - rather than the question - ``did the system emit the right verses somewhere?'' The metric uses a scored region (to exclude instrumental interludes and discussion from scoring), a 1\,s collar at segment boundaries, and gap-tolerant scoring (frames between adjacent segments accept either neighboring line or $\varnothing$). Cold-start offsets are evaluated explicitly, not averaged away.
  \item \textbf{A small, hand-annotated benchmark (v1)}: Initial seed of 4 Kirtan recordings $\times$ 3 cold-start offsets $= 12$ evaluation cases, ${\sim}57$ minutes of scored audio, with a Python scorer, visualizer, and annotation tool.
  \item \textbf{A reference system} demonstrating the benchmark is tractable: a fine-tuned 120M-parameter IndicConformer, a fuzzy line matcher, and a state machine that locks on shabad identity and advances through lines. The full pipeline is INT8 ONNX, runs at RTF ${\approx}0.05$ on a single Apple Silicon CPU core, and scores \textbf{57.9\%} overall (10/12 correct shabad locks) on the difficult variant (live audio, unknown shabad) across all 12 evaluation cases.
\end{enumerate}

This paper offers a \textbf{shared yardstick}: a benchmark, a scorer, and a reference system. We are not alone in working toward live scripture captioning - several efforts in the Sikh tech community target related problems, and in the parallel Quranic tradition Yazin A.'s open-source \texttt{offline-tarteel} (now \texttt{Tilawa}) \citep{offlinetarteel} independently arrived at a closely related ASR $\rightarrow$ matcher $\rightarrow$ state-machine pipeline. For the Sikh context, this work provides shared language to measure accuracy and a deployed reference system as a first baseline.

\section{Task Formalization}

\paragraph{Task.} Given a stream of Kirtan audio, at every time $t$ output a prediction
\[
\hat{y}(t) \in \{(s, \ell) : s \in \mathcal{S},\ \ell \in \mathcal{L}_s\} \cup \{\varnothing\},
\]
where $\mathcal{S}$ is the set of shabads in SGGS and $\mathcal{L}_s$ is the ordered set of lines within shabad $s$. We use the integer identifiers from BaniDB / Sikhi To The Max; the canonical Gurmukhi for any $(s, \ell)$ is recovered via API lookup as an alternative submission path (exact match still required).

\paragraph{Two natural axes.} Systems for this task differ along two binary axes: \emph{temporal access} (offline = full audio available; live = causal, audio up to $t$ only) and \emph{shabad knowledge} (oracle = shabad identity given; blind = must be discovered from audio). The four variants are shown in Table~\ref{tab:matrix}.

\begin{table}[h]
\centering
\small
\begin{tabular}{lll}
\toprule
 & \textbf{Blind} (must identify shabad) & \textbf{Oracle} (shabad given) \\
\midrule
\textbf{Offline} (full audio) & Identify, then align lines & Line alignment only \\
\textbf{Live} (causal) & Identify + follow real-time & Follow lines in real time \\
\bottomrule
\end{tabular}
\caption{The four variants. The same submission format and scorer apply to all four; what differs is the information available to the system.}
\label{tab:matrix}
\end{table}

This paper targets the live+blind variant, but the benchmark scores any variant under the same metric and submission format.

\paragraph{Cold-start variants.} Each recording is additionally evaluated from offsets 33\% and 66\% into the audio, by moving the scored region forward while feeding the same audio. This tests how quickly a system locks on when a listener joins mid-program.

\section{Why Not WER?}

WER and CER are natural metrics for transcription tasks; we deliberately do not use them, for two reasons.

\paragraph{Different layer.} Three layers of representation are relevant:

\begin{table}[h]
\centering
\small
\begin{tabular}{lll}
\toprule
\textbf{Layer} & \textbf{Output} & \textbf{Metric family} \\
\midrule
3. Line identity & $(\shabad, \lineidx)$ over time & \textbf{Frame accuracy (this work)} \\
2. Canonical text & Snap-to-canonical Gurmukhi & Exact-match \\
1. ASR transcript & Noisy Gurmukhi & WER / CER \\
\bottomrule
\end{tabular}
\end{table}

WER measures whether the system transcribed the audio correctly; it is the right metric at the transcription layer, where prior Gurbani ASR work \citep{dua2022} applies it, as do we when evaluating this system's underlying ASR component. The captioning task is a different question: which canonical line is on the projector at time $t$. A system can transcribe perfectly and still display the wrong line (wrong shabad, late by ten seconds), so the benchmark scores the displayed line over time rather than transcription fidelity.

\paragraph{Different cultural constraint.} In Kirtan, misspelled Gurmukhi is religiously inappropriate. A canonical-only metric forces systems to either output real Gurbani or output nothing - which matches what the product can ethically display.

\section{The v1 Benchmark}

\paragraph{Data.} Four publicly available Kirtan recordings on YouTube were selected as v1, chosen for clean audio and the simplest task structure (one shabad per recording, no Katha, Simran, or interludes). Each recording yields three evaluation cases: the full file, and cold-start variants at 33\% and 66\%. Recordings range from 4.9 to 10.9 minutes; this initial sample is ${\sim}57$ minutes of scored audio across 12 cases (Table~\ref{tab:data}).

\begin{table}[h]
\centering
\small
\begin{tabular}{lrrr}
\toprule
\texttt{video\_id} & \texttt{shabad\_id} & Duration & Lines \\
\midrule
\texttt{IZOsmkdmmcg} & 4377 & 7.7 min & 16 \\
\texttt{kZhIA8P6xWI} & 1821 & 5.1 min & 19 \\
\texttt{kchMJPK9Axs} & 1341 & 10.9 min & 22 \\
\texttt{zOtIpxMT9hU} & 3712 & 4.9 min & 10 \\
\bottomrule
\end{tabular}
\caption{The four base recordings in v1. Each yields three evaluation cases via cold-start offsets.}
\label{tab:data}
\end{table}

\paragraph{Ground-truth format.} Each case is one JSON file containing the \texttt{shabad\_id}, total duration, a scored region \texttt{\{start, end\}} (frames outside it are unscored - this excludes intros, outros, and the cold-start skip region), and a list of \texttt{segments}, each a \texttt{(start, end, line\_idx)} triple. \texttt{line\_idx} is 0-indexed within the shabad. The canonical Gurmukhi text of each line is included inline in the GT file so submissions using the text-key path (\texttt{banidb\_gurmukhi}) can be scored without an API lookup.

\paragraph{Submission format.} A submission is one JSON file per GT case, listing predicted segments \texttt{(start, end, line\_idx)}. We ship a few different paths for producing this JSON - direct file write, a relay that impersonates the STTM display protocol for systems that already drive it, and a minimal client adapter - but all produce the same final format, which is what the scorer reads.

\paragraph{Scoring.} Time is discretized to 1-second frames; every frame within the scored region is evaluated. Each frame falls into one of three regions (Table~\ref{tab:scoring}).

\begin{table}[h]
\centering
\small
\begin{tabular}{>{\raggedright\arraybackslash}p{0.18\linewidth}>{\raggedright\arraybackslash}p{0.38\linewidth}>{\raggedright\arraybackslash}p{0.34\linewidth}}
\toprule
\textbf{Region} & \textbf{Definition} & \textbf{Accepted predictions} \\
\midrule
Segment interior & Inside a labeled segment, $>$collar from either edge & Exact \texttt{line\_idx} only \\
Collar & Within collar seconds of a segment boundary & Exact line, adjacent line, or $\varnothing$ \\
Gap & Between two adjacent segments, outside both collars & Line before gap, line after gap, or $\varnothing$ \\
\bottomrule
\end{tabular}
\caption{Frame-region rules in the scorer. Frames outside any segment and outside any inter-segment gap (e.g., before the first segment) are unscored.}
\label{tab:scoring}
\end{table}

The primary number reported is \textbf{frame accuracy at collar $= 1$\,s}. The collar credits systems that handle line transitions within $1$\,s of the true boundary; tolerable latency is itself a property of the deployment, so the collar is set tight enough to discriminate real-time-capable systems from sluggish ones, but not so tight that small timing differences dominate the score.

\paragraph{Baselines.}

\begin{table}[h]
\centering
\small
\begin{tabular}{lll}
\toprule
\textbf{Baseline} & \textbf{Description} & \textbf{Frame accuracy} \\
\midrule
\texttt{empty} & No segments emitted - $\varnothing$ everywhere & 26.0\% \\
\texttt{shifted\_5s} & Ground truth, every segment delayed by 5 seconds & 85.5\% \\
\texttt{perfect} & Exact copy of ground truth & 100.0\% \\
\bottomrule
\end{tabular}
\caption{Reference baselines on v1.}
\label{tab:baselines}
\end{table}

The \texttt{empty} baseline is non-trivially above 0\% because inter-segment gaps accept $\varnothing$ as correct; it functions as a sanity check on the scorer. The \texttt{shifted\_5s} baseline approximates the latency profile of an online ASR pipeline that lags audio by a few seconds - the 1\,s collar absorbs a small fraction; the rest of the cost shows up in segment interiors. Confidently emitting the \emph{wrong} shabad scores worse than emitting nothing, by design.

\begin{figure}[!t]
\centering
\includegraphics[width=\linewidth]{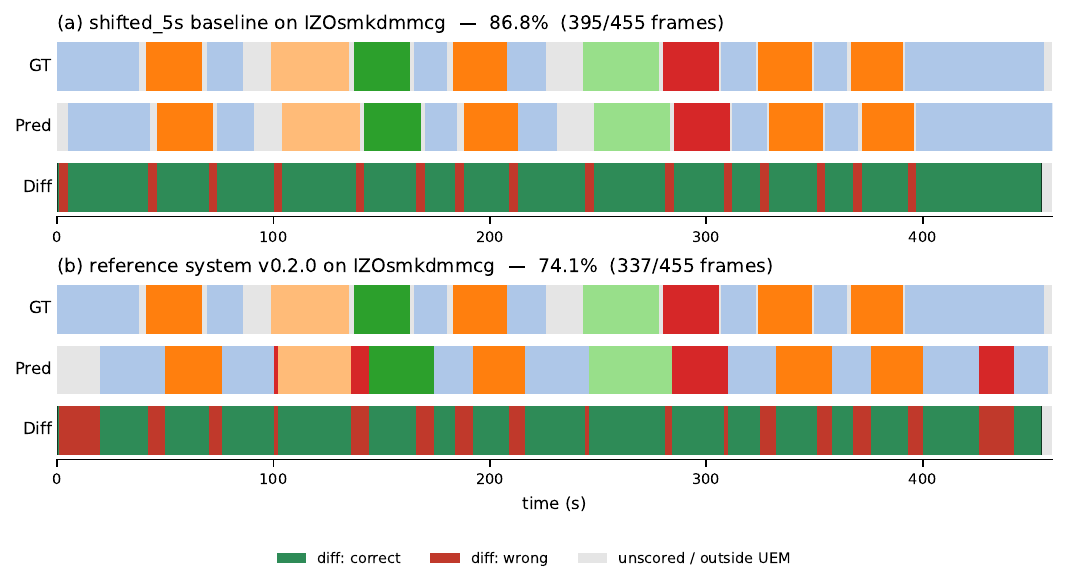}
\caption{\textbf{Frame-accuracy visualization on \texttt{IZOsmkdmmcg}.} Three strips per panel: ground-truth line track (GT), system prediction (Pred), and per-frame diff (green = correct under collar/gap rules, red = wrong, faint = unscored / outside the scored region). Scored-region boundaries are marked by hairlines on the diff strip. (a) The \texttt{shifted\_5s} baseline: predictions trail GT by 5\,s; the 1\,s collar absorbs a small slice of each transition, the remainder is paid as red in segment interiors. (b) The reference system on the same recording: locks correctly to S4377 at $t=18$\,s and tracks lines from there onward; the red prefix is the pre-lock region, the green body is correct in-shabad tracking. Same metric exposes ``latency cost'' and ``cold-start cost'' as distinct quantities.}
\label{fig:strips}
\end{figure}

\paragraph{Implementation.} The scorer (\texttt{eval.py}) is standard-library Python 3.10+. A companion script, \texttt{visualize.py}, renders an interactive HTML page with GT/prediction/diff strips, audio scrubbing, hover-tooltips showing canonical Gurmukhi, and a cold-start toggle.

\section{Reference System}

We describe a reference system targeting the hardest variant (live $\times$ blind), to demonstrate the benchmark is tractable. The system runs as a three-stage pipeline.

\paragraph{Stage 1: ASR.} We fine-tuned the AI4Bharat 120M-parameter Punjabi IndicConformer (Hybrid CTC-RNNT) \citep{indicconformer2025} on weakly-supervised Kirtan clips. Two pipelines feed the training set, both of which align a noisy transcript to the canonical SGGS text and keep only high-confidence lines: (i) starting from existing transcripts (subtitles, larger ASR outputs, or earlier model outputs), and (ii) generating fresh transcripts with Google Chirp, then identifying the shabad over a window of audio and snapping each transcript word to the closest canonical line within it. In both cases the canonical text serves as the cleanup target: noisy words near a canonical line word are corrected; lines below a confidence threshold are dropped. The fine-tuned model is exported to \textbf{INT8 ONNX}; on a single Apple Silicon CPU core, a 10-second window decodes in ${\sim}490$\,ms (RTF ${\approx}0.05$); on a commodity cloud vCPU we measured RTF ${\approx}0.08$. At inference time, once a shabad is locked (Stage 3 below), CTC decoding is hard-constrained to a trie of that shabad's lexicon: the beam search can only emit token sequences that spell words from the locked shabad, which can improve line-level accuracy once the shabad is correctly locked. The unconstrained decoder is still used during the blind cold-start phase.

\paragraph{Stage 2: Matcher.} The matcher runs in two phases. \textbf{Phase 1 (identification)} scores a noisy Gurmukhi transcript $T$ against the entire SGGS text by fuzzy string matching against every line. For each candidate shabad $s$, we compute
\[
\textsf{score}(T, s) \;=\; \alpha\,\max_{\ell \in \mathcal{L}_s} \textsf{sim}(T, \ell) \;+\; \beta\,\min(g_s,\, 5),
\]
where $\textsf{sim}(\cdot,\cdot)\in[0,100]$ is a fuzzy substring-similarity score (RapidFuzz partial\_ratio), $g_s = |\{\ell \in \mathcal{L}_s : \textsf{sim}(T,\ell) \ge \tau\}|$ is the number of the shabad's lines that match $T$ above a threshold $\tau$, and $(\alpha, \beta) = (0.7, 6)$ are tuned weights. The first term rewards the single best-matching line; the second is a multi-line agreement bonus, capped at five lines so that a few strongly matching lines suffice. Shabads with very few lines receive no agreement bonus, since they cannot demonstrate agreement. \textbf{Phase 2 (line tracking)} runs only after a shabad is locked (Stage 3): it scores the transcript against the ${\sim}10{-}20$ lines of the locked shabad alone, using a bidirectional fuzzy word-overlap score (the F1 of line-word recall and window-word precision). Like the hard-CTC decoder, this scorer is restricted to the locked shabad and exploits the lock; until lock, only Phase 1 runs.

\paragraph{Stage 3: State machine.} A small state machine consumes matcher outputs on two cadences: an identification tick over a wider window (${\sim}30$\,s) when not yet locked, and a tracking tick over a shorter window (${\sim}15$\,s) once locked. In autolock mode (used for the benchmark), the lock decision fires when several gates pass together: a minimum amount of audio has elapsed, the top-ranked shabad has stayed top-ranked for several consecutive identification windows, and the relative score gap between top-1 and top-2 candidates exceeds a confidence threshold (with a more permissive threshold available later in the audio, so a system that has heard a lot but is only moderately confident can still commit). The thresholds are tuned hyperparameters. Once locked, a within-shabad pointer advances through lines biased to the matcher's best line per chunk; a small \textbf{hysteresis margin} on line switching prevents per-window display flicker, requiring a competing line to beat the displayed one by a margin before it takes over. The pointer does not assume monotonic forward progress: Kirtan repeatedly returns to the \emph{rahao} (the recurring refrain that alternates with verses), and a strictly forward tracker would treat legitimate \emph{rahao} returns as backward jumps and suppress them. A recovery mechanism guards against bad locks: an \textbf{auto-unlock sliding window} tracks recent line-match scores and falls back to identification if they sustain below a dynamic threshold. An optional periodic \textbf{sanity check} can additionally re-run full identification and force unlock if the locked shabad falls out of the top-$K$ candidates.

\paragraph{Deployment.} The full system has a live deployment (current demo link in the reference-system repository) that continuously captions Sikhnet Radio. It is also deployed as a desktop app in the GitHub repository, and as an on-device iOS app (a Swift port of the same engine, INT8 ONNX / CoreML), distributed via a TestFlight beta linked from the reference-system repository. The autonomous machinery described above - multi-gate autolock, hysteresis, auto-unlock - is what the benchmark exercises: the scorer requires the system to make committed decisions over time with no human in the loop. The deployment runs a different UX on top of the same underlying model: the interface displays top-$k$ shabad candidates rather than autonomously committing, and the listener can confirm a candidate or reset the tracker. The benchmark numbers in Section~\ref{sec:results} therefore reflect the system deciding alone, with no human in the loop; the assisted deployment, where a listener confirms candidates, should do at least as well. We view assisted-commit as the appropriate experience today - sewadars must retain control over the display, especially in group settings.

\section{Results}
\label{sec:results}

We report frame accuracy on all 12 evaluation cases (4 recordings $\times$ 3 cold-start offsets) on the hardest variant (live $\times$ blind). Results are from \texttt{kirtan-captioning} v0.2.0 (commit \texttt{3b42245}), model \texttt{v4.int8.onnx}, scored with \texttt{eval.py {-}{-}collar 1} on macOS arm64 (CPUExecutionProvider). Predictions are released alongside this paper for direct re-scoring.

\begin{table}[h]
\centering
\small
\begin{tabular}{lrrl}
\toprule
\textbf{Case} & \textbf{Frame acc.} & \textbf{Frames} & \textbf{Shabad lock} \\
\midrule
\texttt{kchMJPK9Axs}         & 82.0\% & 532/649 & \checkmark\ S1341 @ 23s \\
\texttt{kchMJPK9Axs\_cold33} & 80.1\% & 351/438 & \checkmark\ S1341 @ 259s \\
\texttt{kchMJPK9Axs\_cold66} & 73.0\% & 162/222 & \checkmark\ S1341 @ 475s \\
\texttt{IZOsmkdmmcg}         & 74.1\% & 337/455 & \checkmark\ S4377 @ 18s \\
\texttt{IZOsmkdmmcg\_cold33} & 10.7\% & 33/308  & $\times$\ locked S3643 (wrong) \\
\texttt{IZOsmkdmmcg\_cold66} & 68.0\% & 106/156 & \checkmark\ S4377 @ 322s \\
\texttt{kZhIA8P6xWI}         & 58.1\% & 176/303 & \checkmark\ S1821 @ 33s \\
\texttt{kZhIA8P6xWI\_cold33} &  3.9\% & 8/207   & $\times$\ locked S24 (wrong) \\
\texttt{kZhIA8P6xWI\_cold66} & 44.8\% & 47/105  & \checkmark\ S1821 @ 242s \\
\texttt{zOtIpxMT9hU}         & 30.3\% & 87/287  & \checkmark\ S3712 @ 188s \\
\texttt{zOtIpxMT9hU\_cold33} & 49.0\% & 96/196  & \checkmark\ S3712 @ 174s \\
\texttt{zOtIpxMT9hU\_cold66} & 47.5\% & 47/99   & \checkmark\ S3712 @ 216s \\
\midrule
\textbf{Overall}             & \textbf{57.9\%} & \textbf{1982/3425} & \textbf{10/12} \\
\bottomrule
\end{tabular}
\caption{Per-case results on v1 (live $\times$ blind variant), reference system v0.2.0.}
\label{tab:results}
\end{table}

\paragraph{Headline.} Overall frame accuracy is \textbf{57.9\%} (1{,}982 of 3{,}425 scored frames) with \textbf{10 of 12 correct shabad locks}. Both lock failures are \texttt{\_cold33} cases - dropped into the middle of a recording with no prior audio history, the matcher's first 30\,s of audio does not contain enough distinctive words to separate the target shabad from confusable candidates. This is the hardest variant in the suite.

\begin{figure}[H]
\centering
\includegraphics[width=\linewidth]{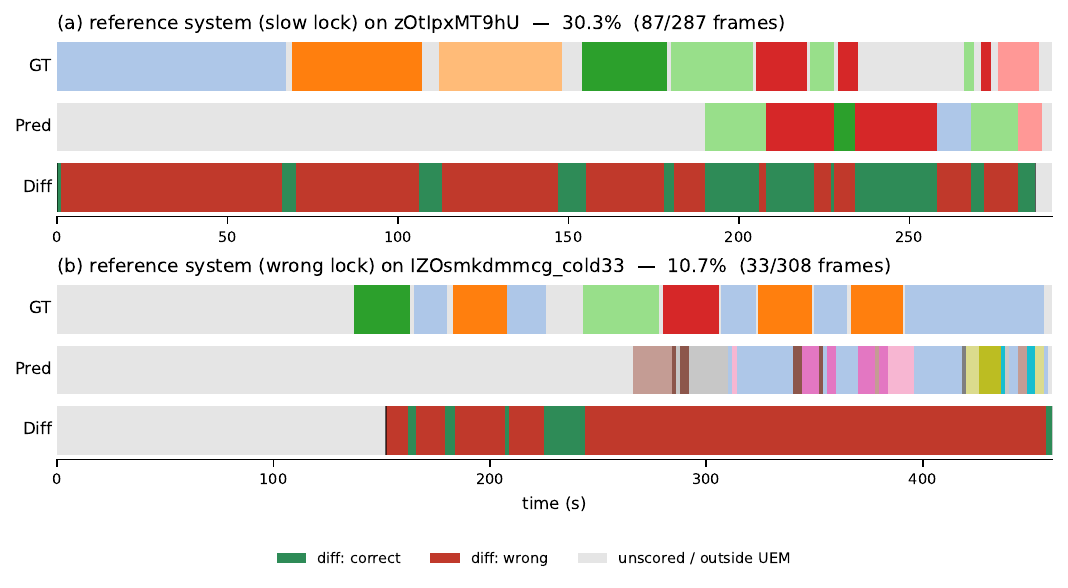}
\caption{\textbf{Two failure modes the frame-temporal metric distinguishes.} (a) \emph{Slow lock, then good tracking}: \texttt{zOtIpxMT9hU} locks correctly to S3712 at $t=188$\,s; the diff strip is red until lock and green afterward. (b) \emph{Confidently wrong lock}: \texttt{IZOsmkdmmcg\_cold33} commits to the wrong shabad (S3643 instead of S4377) at $t=260$\,s and then bounces rapidly between 10 of S3643's lines (the busy multi-color Pred strip), so every interior frame fails the exact-line rule and the diff strip is solid red. An utterance-level retrieval metric would conflate these two outcomes; the frame-temporal metric separates them.}
\label{fig:failures}
\end{figure}

\paragraph{Where the cost is paid.} Inspecting the per-case lock timestamps shows two distinct failure modes that the frame-temporal metric exposes but which utterance-level retrieval metrics would conflate (Figure~\ref{fig:failures}):

\begin{enumerate}[leftmargin=*,itemsep=2pt]
  \item \textbf{Slow lock, then good tracking.} \texttt{zOtIpxMT9hU} locks correctly but only at 188\,s (more than half the recording is gone), so the score is dominated by the pre-lock red region rather than by tracking errors. The diff strip is red for the first ${\sim}3$ minutes and green thereafter.
  \item \textbf{Confidently wrong lock.} \texttt{IZOsmkdmmcg\_cold33} locks to the wrong shabad (S3643 instead of S4377) and stays wrong; this scores worse than the empty baseline on this case because frames in segment interiors only accept the exact \texttt{line\_idx}.
\end{enumerate}

The \texttt{shifted\_5s} baseline scores 85.5\%, so there remains a substantial gap between the reference system and an oracle pipeline with only latency error - most of the gap comes from shabad identification (the two wrong locks plus the several slow locks) rather than within-shabad tracking, which is reasonably accurate once a lock is established.

\section{Related Work}

\paragraph{Parallel work on Quranic recitation.} Yazin A.'s open-source \texttt{offline-tarteel} (since renamed \texttt{Tilawa}) \citep{offlinetarteel} is an on-device system that identifies the Quranic verse being recited from audio with a similar ASR $\rightarrow$ fuzzy matcher $\rightarrow$ state-machine pipeline. It targets a different scripture and reports verse-retrieval metrics. Separately, and despite the shared name, Tarteel Inc.\ \citep{tarteelinc} is an unrelated closed-source Quran-memorization product that tracks the user's place within a chosen verse, an oracle-variant analogue at finer (word-level) granularity.

\paragraph{Direct prior Gurbani ASR.} \citet{dua2022} build a CNN speech-to-text system for Gurbani hymns and report WER; that measures transcription, whereas we measure which canonical line is displayed over time.

\paragraph{Generic Punjabi ASR.} A broader research community works on general-purpose Punjabi ASR without addressing Kirtan or the canonical-output constraint.

\section{Limitations and Future Work}

\paragraph{Limitations of v1.}
\begin{itemize}[leftmargin=*,itemsep=2pt]
  \item \textbf{Scale.} Four recordings is small; statistical claims about system comparisons would be noisy. We will extend to more recordings.
  \item \textbf{Task simplicity.} v1 recordings contain one shabad each, with no Katha (spoken commentary), Sehaj Paath (a complete reading of the SGGS at a relaxed pace), Simran (meditative repetition), or interludes. These are common in real Gurdwara programs and a future update will include them.
  \item \textbf{SGGS only.} \emph{Dasam Bani} and other Sikh scripture are not yet included.
\end{itemize}

\paragraph{Limitations of the reference system.}
\begin{itemize}[leftmargin=*,itemsep=2pt]
  \item Sensitive to recitation speed (very slow or very fast hurts the matcher).
  \item No awareness of adjacent shabads on the same \emph{ang} (page), so Akhand Path / Nitnem flows are not tracked smoothly. An update with support for this continuous recitation is in testing.
  \item Degrades on low-quality audio; opportunity to fine-tune with intentionally noisy data.
\end{itemize}

\paragraph{Future work.} STTM-instrumented capture (using deployed sessions as weak ground truth) is an attractive route to scale annotation; we ship a prototype recorder (\texttt{sttm\_recorder.py}) that could serve as the data-collection endpoint. A related direction is to measure and tune the training-data pipeline of Section~5 as a labeling loop in its own right - how often does retrieval pick the wrong shabad, what threshold settings trade label quality for label volume, and how does that trade-off affect downstream model quality - which could be a separate data-centric paper.

The reference system has many directions worth exploring: alternative ASR models specialised for Kirtan (recent public examples include \citealp{surindersingh2026kirtanmodel}), alternative matcher designs, and replacing parts of the hand-tuned state machine with learned components. A complementary direction is normalising the implementation of Yazin's \texttt{offline-tarteel}/\texttt{Tilawa} \citep{offlinetarteel} against ours - both pipelines decompose into ASR, matcher, and state-machine layers, so an apples-to-apples comparison would let the best ideas from each tradition feed back into the other. The system can also serve as a baseline reference for other Gurbani applications to include related functionality. Lastly, we hope to leverage this work to remove inaccuracies in current automated video captioning platforms that treat Gurbani no different than generic speech.

\section{Ethics and Cultural Considerations}

The choice to enforce a canonical-only output is not technical. Misspelled Gurmukhi in a religious display context is \emph{not a neutral error}; it is read as disrespect to the Guru Granth Sahib Ji, revered by Sikhs as a living Guru. We therefore designed both the metric and the system so that emitting nothing is preferable to emitting the wrong text, and we argue against any deployment of free-form ASR output as the display layer in this context.

Our current deployment runs the reference system in beta with a confirm-style interface - the system shows candidates and a listener taps to commit - rather than autonomously driving the display.

The pipeline is built around a small on-device model that can run on consumer hardware. Users do not need to send their audio to any external servers for processing; the iOS app runs the model entirely on-device.

\section{Availability}

\begin{itemize}[leftmargin=*,itemsep=2pt]
  \item Benchmark: \url{https://github.com/karanbirsingh/live-gurbani-captioning-benchmark-v1} (code MIT, annotations CC BY 4.0)
  \item Reference system: \url{https://github.com/karanbirsingh/kirtan-captioning} (CC BY-NC-SA 4.0)
  \item iOS app (on-device, INT8 ONNX / CoreML): TestFlight beta linked from the reference-system repository \texttt{README}
  \item Live deployment: current demo link in the reference-system repository \texttt{README}
  \item Visualizations: \url{https://karanbirsingh.github.io/live-gurbani-captioning-benchmark-v1}
\end{itemize}

\paragraph{Reproducibility.} The scorer (\texttt{eval.py}) is standard-library Python 3.10+. The exact prediction JSONs that produced Table~\ref{tab:results} are committed under \texttt{benchmark/results/} in the reference-system repository, alongside the model checksum and a single-command reproduction recipe.

\section*{Acknowledgments}

Thanks to friends from within and outside the Sikh community who served as a sounding board throughout this work, to the maintainers of BaniDB and Sikhi To The Max for open canonical-text APIs, and to the reciters whose recordings comprise v1.

\bibliographystyle{plainnat}
\bibliography{refs}

\end{document}